\documentclass[preprint,review,12pt]{elsarticle}
\usepackage{amssymb}
\usepackage{amsthm}
\usepackage[figuresright]{rotating}
\usepackage{amsmath}
\usepackage{mathrsfs}
\usepackage{newtxmath}
\usepackage{color, colortbl}
\usepackage{amsfonts}
\usepackage{mathtools}
\usepackage{bbding}
\usepackage{algorithm}
\usepackage{algorithmic}
\usepackage{soul}
\usepackage{url}
\usepackage[utf8]{inputenc}
\usepackage{caption}
\usepackage{subcaption}
\usepackage{graphicx}
\usepackage{booktabs}
\usepackage{xcolor}
\usepackage{multirow}
\usepackage{verbatim}
\usepackage{float}
\usepackage{textcomp}
\usepackage{lipsum}
\usepackage{wrapfig}
\usepackage{stfloats}
\usepackage{array}
\usepackage{comment}
\usepackage{makecell}

\usepackage{setspace} 

\usepackage{hyperref}
\hypersetup{
    colorlinks=true,
    linkcolor=blue,
}
\journal{Pattern Recognition}
\begin{document}

\begin{frontmatter}

\title{Super Encoding Network: Recursive Association of Multi-Modal Encoders for Video Understanding}

\author[inst1,inst2]{Boyu Chen\corref{cor1}}
\cortext[cor1]{Equal Contribution}
\author[inst1,inst2]{Siran Chen\corref{cor1}}
\author[inst1,inst2,inst3]{Kunchang Li}
\author[inst1,inst2]{Qinglin Xu}
\author[inst3]{Yu Qiao}
\author[inst1,inst3]{Yali Wang\corref{cor2}}
\cortext[cor2]{Corresponding author.}

\affiliation[inst1]{organization={Shenzhen Institute of Advanced Technology, Chinese Academy of Sciences},
            city={Shenzhen},
            country={China}}
\affiliation[inst2]{organization={the School of Artificial Intelligence, University of Chinese Academy of Sciences},
            city={Beijing},
            country={China}}
\affiliation[inst3]{organization={Shanghai AI Laboratory},
            city={Shanghai},
            country={China}}

\begin{abstract}
Video understanding has been considered as one critical step towards world modeling, which is an important long-term problem in AI research.
Recently, multimodal foundation models have shown such potential via large-scale pretraining. \textcolor{black}{These models effectively align encoders of different modalities via contrastive learning. To further enhance performance on complex target movements and diversified video scenes, we propose to augment this alignment with deeper multimodal interactions}, which are critical for understanding complex target movements with diversified video scenes. To fill this gap, we propose a unified Super Encoding Network (SEN) for video understanding, which builds up such distinct interactions through the recursive association of multimodal encoders in the foundation models. Specifically, we creatively treat those well-trained encoders as ``super neurons" in our SEN. Via designing a Recursive Association (RA) block, we progressively fuse multi-modalities with the input video, 
based on knowledge integrating, distributing, and prompting of super neurons in a recursive manner. In this way, our SEN can effectively encode deeper multimodal interactions for prompting various video understanding tasks in the downstream. Extensive experiments show that our SEN can remarkably boost the four most representative video tasks, including tracking, recognition, chatting, and editing, e.g., for pixel-level tracking, the average jaccard index improves 2.7\%, and temporal coherence(TC) drops by 8.8\% compared to the popular CaDeX++ approach. For one-shot video editing, textual alignment improves 6.4\%, and frame consistency increases by 4.1\% compared to the Tune-A-Video approach.
\end{abstract}


\end{frontmatter}

\section{Introduction}
Due to the ever-increasing volume of model parameters and training text corpora, 
Large Language Models (LLMs)~\cite{gpt4} have demonstrated remarkable performance in a variety of natural language tasks.
Even though language is good at high-level abstraction, it is not always sufficient to capture low-level dynamics in the physical world~\cite{world_model}.
Recent studies have shown that video is a promising ``language" to describe such underlying details, since it records a wealth of spatial-temporal movements within various complex scenarios in our world~\cite{uniformerv2, videomae}. Hence, understanding what happens in the video is an important problem in the multimedia and computer vision community, and it has been seen as a key step for world modeling~\cite{world_model, chen2025g, chen2025super,chen2022low}.

\begin{figure*}[t]
% \vspace {-2em}
\begin{center}
\includegraphics[width=0.9\linewidth]{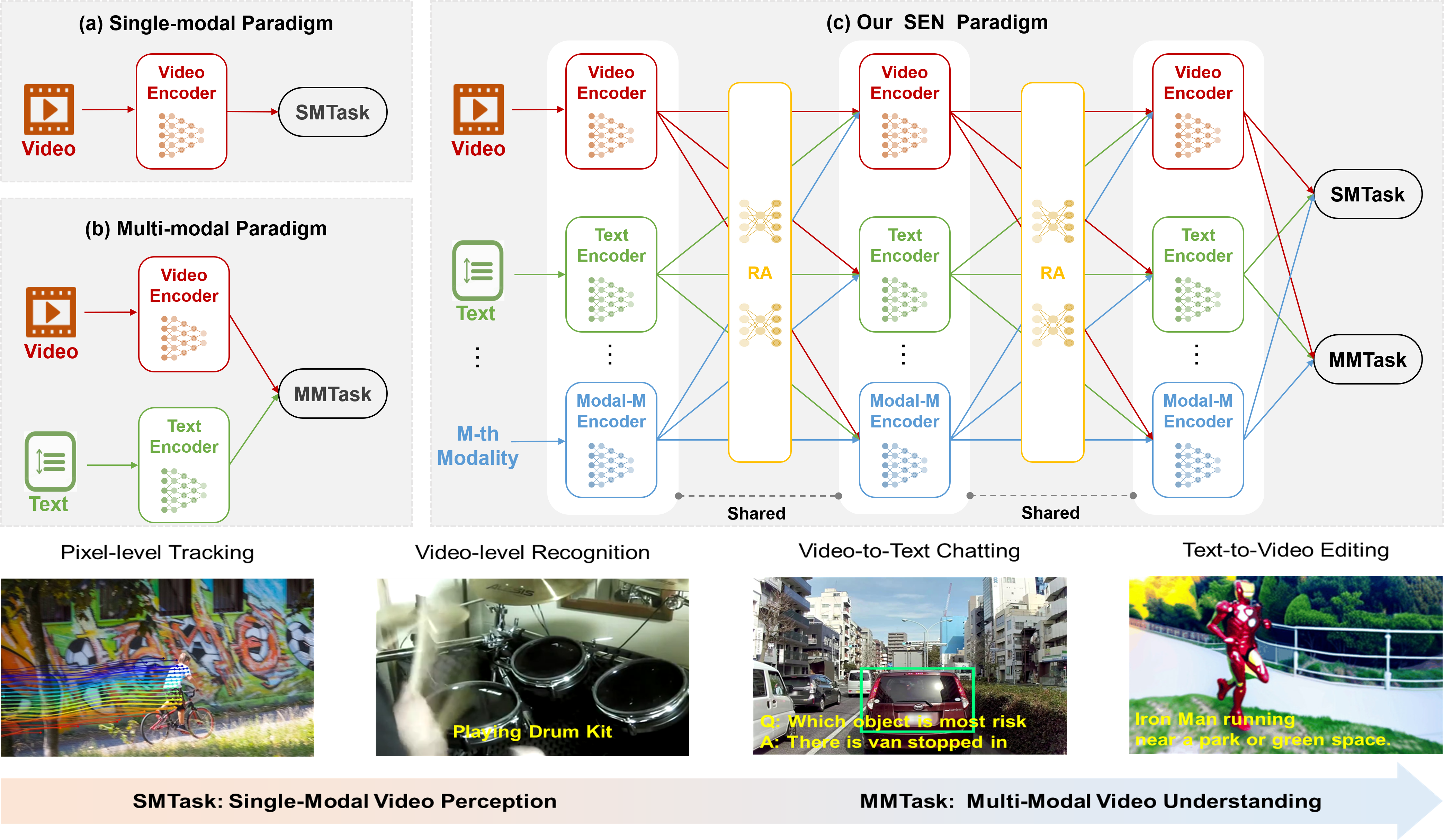}
\end{center}
\vspace {-2em}
\caption{Our Motivation. 
Different from the previous paradigms,
our SEN
creatively treats the well-pretrained multimodal encoders as super neurons,
and
leverages a novel Recursive Association (RA) block to achieve deeper multimodal interaction of super neurons. 
Such a flexible paradigm enables it to serve as a unified encoder network for boosting various complex downstream video understanding tasks.
}
\label{fig:overview}
\vspace{-1em}
\end{figure*}

Traditionally, video understanding mainly works on designing expert models for single-modal video perception tasks~\cite{videomae},
as shown in Fig~\ref{fig:overview}. 
However, our real world naturally consists of multiple modalities.
Apparently, such single-modal designs restrict model generality for capturing multiple modalities in videos.
To fill this gap, recent research has gradually focused on multimodal video understanding, such as video dialogue and generation~\cite{video-llama}, 
based on the fast development of foundation models~\cite{internvideo2, chen2025top, chen2025vragent, chen2025videochat, wang2025videochat}.
\textcolor{black}{While these foundation models effectively align encoders of various modalities via cross-modal contrastive learning, augmenting this global alignment with deeper multimodal interactions can further enhance the capability of capturing,
which can be critical for understanding complex targets in a diversified video scene~\cite{vlmo, albef}.}
For example, it is difficult for models to understand some specific details in language, such as quantitative relationships and object characteristics.  
With deeper multimodal interaction, the model can better explore multimodal information, enhance its robustness, and achieve a better understanding of multimodal data in different tasks.

% However, without more deeper engagement with the text and video content, the model fails to discern
% fine-grained details within the edit prompt leading to inaccuracies.
% \textcolor{red}{XXX} %%% 可以举个例子。
%The deliberative and logical 'slow thinking'~\cite{kahneman2011thinking} process is critical for understanding confusing data and diversified motions in complex and open dynamical scenes.
Alternatively, the human brain can effectively capture such deeper interaction via multimodal association~\cite{lansner2009associative,associativenature}.
After one modality stimulation,
one corresponding brain area will produce neurotransmitters.
Next, the neurotransmitters of this modality are delivered to other brain areas that refer to other modalities. 
This procedure is recursively repeated many times to achieve deep association among modalities for better understanding. 
%can be regardes as a multimodal 'slow thinking' process 
%Different from the Chain of Thought~\cite{cot}, this multimodal 'slow thinking' process has intermediate thinking processes in the form of compressed multimodal knowledge like 'latent' in the deep-learning network, which can not be formulated by natural language~\cite{coconut}.

Inspired by these facts,
we propose a unified and concise Super Encoding Network (SEN) for video understanding.
Different from the previous multimodal foundation models,
our SEN is a meta encoder network,
where we creatively leverage multimodal encoders as super neurons, as shown in Fig~\ref{fig:overview}.
More importantly, \textcolor{black}{we design a novel lightweight Recursive Association (RA) block for task adaptation.}
Via knowledge integrating, distributing, and prompting of the super neurons in a recursive manner,
our SEN can effectively encode deep interactions of multi-modalities within multi-layer stacking,
for boosting various complex video tasks in downstream scenarios.
To show the effectiveness of our SEN, we chose the four most representative video tasks 
including  
pixel-level tracking, 
video-level recognition,
video-to-text chatting,
text-to-video editing,
which range from single-modal perception to multimodal understanding,
from full video supervision to few-shot video learning.
Extensive experiments show that our SEN significantly improves the performance across these tasks. 
In the pixel-level tracking task, the temporal coherence (TC) declines 8.8\%, with the value from 0.68 to 0.62.
In the video-level recognition task, Top1 accuracy increases 2.1\%.
In the video-to-text chatting task, the CIDER index shows a 5.0\% improvement, with the value rising from 310.1 to 325.5. 
In the text-to-video editing task, the textual alignment increases by 6.4\%. 

\section{Related work}
\textbf{Video Understanding.}
\textcolor{black}{
Early approaches relied on CNN architectures, employing dual-branch designs~\cite{slowfast} or decomposed convolutions~\cite{tsn} for spatiotemporal modeling. 
Recently, Video Transformers~\cite{videomae} have leveraged pre-trained ViT~\cite{vit} representations, significantly advancing performance across diverse tasks, including action recognition~\cite{liu2025sam,wang2025vlpa}, generation~\cite{tune-a-video, cogvideo}, and question answering~\cite{cai2025mllm}. 
However, most traditional methods operate as specialized expert models tailored to single task domains~\cite{uniformerv2}. 
Consequently, they often lack the architectural flexibility to generalize across the broad spectrum of multimodal video understanding tasks. }

\textbf{Multi-modal Foundation Models.}
\textcolor{black}{
Recent research has rapidly advanced by aligning diverse modalities via contrastive learning.
Pioneered by CLIP~\cite{clip} on image-text pairs, this paradigm has been extended by VALOR~\cite{valor} and ImageBind~\cite{imagebind} to incorporate audio, depth, and thermal modalities.
While scaling models enhances multimodal comprehension~\cite{videomae,internvideo2}, they often require prohibitive computational cost for retraining and focus primarily on global alignment.
In contrast, our approach functions as a parameter-efficient pipeline. 
By freezing multimodal encoders, we leverage their pre-trained knowledge while introducing recursive associations to achieve deeper multimodal interaction with minimal computational cost.}

\textbf{Knowledge Transfer Learning for Downstream Tasks: } 
\textcolor{black}{To adapt foundation models, prior works utilize parameter-efficient techniques like prompt learning~\cite{Cocoop,CoOp} and adapters~\cite{clipadapter}, or employ agent-based collaboration~\cite{chen2025percept}.
While effective, these paradigms predominantly rely on the pre-established global alignment of encoders.
However, relying solely on global alignment may limit the potential for capturing intricate fine-grained dynamics. 
By augmenting this with deeper multimodal interaction, we aim to further boost the generalization and modeling capability of these foundation models in open-world scenarios.}
% However, without deeper multimodal interaction, these foundation models lack generalization and modeling capability.

\begin{figure*}[t]
\begin{center}
% \vspace{-2em}

\includegraphics[width=\textwidth]{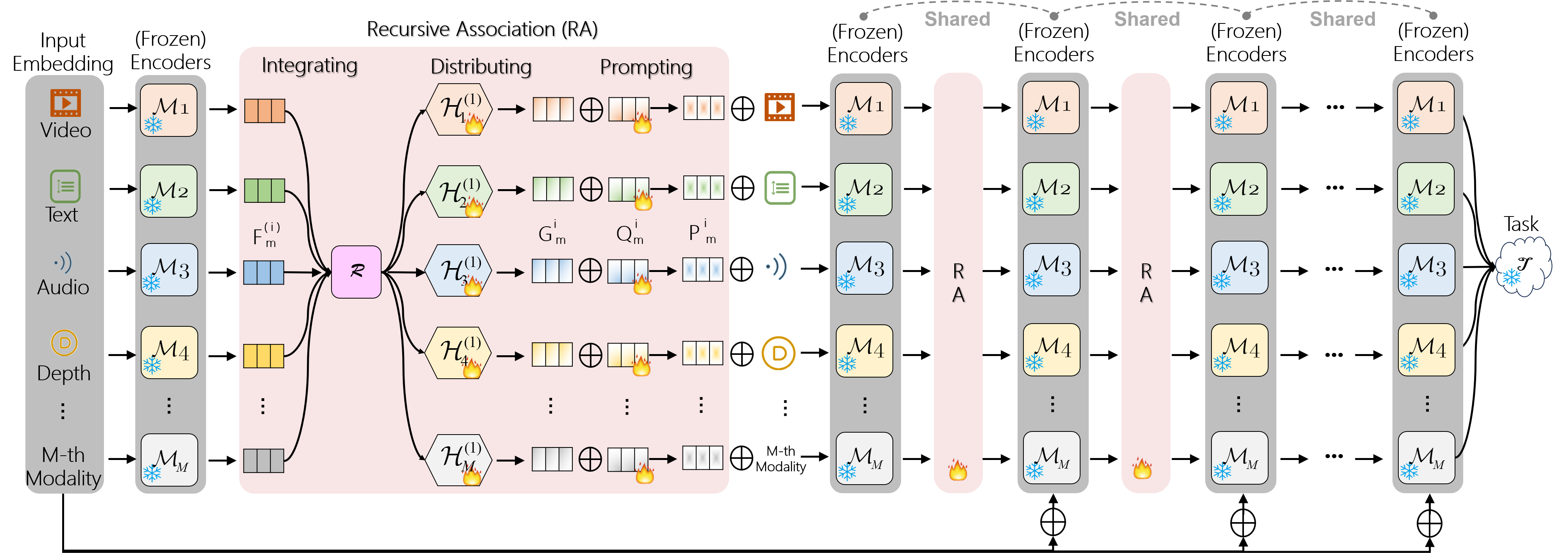}
\end{center}
\vspace{-1.5em}
\caption{\textcolor{black}{ Overall Structure of Our SEN.
Each neuron is one of the multimodal encoders in the pretrained foundation model.
We design a Recursive Association (RA) block to learn deeper interactions of multimodal knowledge in a progressive manner.
Subsequently, we utilize the final feature of SEN to boost various complex video tasks in the downstream.}
}
\vspace{-1em}
\label{fig:main structure}
\end{figure*}

\section{Method}
\subsection{Our Super Encoding Network (SEN)}

In this section,
we introduce our SEN in detail.
As shown in Fig~\ref{fig:main structure},
there are two distinct characteristics.
First, it is a meta-encoder network, since each neuron refers to an encoder network of one modality that comes from a multimodal foundation model.
This allows it to inherit the well-pretrained knowledge of these frozen foundation models.
Second, it is a recursive structure, where we cascade these super neurons via a Recursive Association (RA) block.
This enables it to \textcolor{black}{achieve a substantial Interaction Depth ($D$)}, thereby facilitating the learning of intricate cross-modal dependencies critical for enhancing various complex video tasks in the downstream.
%%%%%%%%%%%%%%%%%%%%%%% Figure 2

As shown in Fig~\ref{fig:main structure}, input multimodal data $\{\mathbf{X}_{m}\}_{m=1}^{M}$ are first fed into the first super neuron layer to extract each modality's latent feature.
\begin{equation}
\setlength\abovedisplayskip{1pt}
\setlength\belowdisplayskip{1pt}
\mathbf{F}^{(1)}_{m}=\mathcal{M}_{m}(\mathbf{X}_{m}).
\label{eq:embedding}
\end{equation}
where $\mathbf{F}^{(1)}_{m}$ is the latent feature of modality $m$ at the \textcolor{black}{1st recursive step}.
Notice that $\mathcal{M}_{m}$ represents the encoders in the frozen foundation models, \textcolor{black}{e.g., the 32-layer Vision Transformer in ImageBind~\cite{imagebind}}.
Hence, the latent features $\{\mathbf{F}^{(1)}_{m}\}_{m=1}^{M}$ inherit semantic knowledge of multimodalities,
which are memorized in the frozen foundation models via large-scale data pre-training.
\textcolor{black}{While these models effectively align encoders via contrastive learning, augmenting this alignment with deeper multimodal interactions is essential for modeling complex fine-grained dynamics in open scenarios.}
To this end, 
we propose a concise Recursive Association (RA) block,
which \textcolor{black}{establishes a feedback loop that can} progressively fuse multi-modalities by knowledge integrating, distributing, and prompting \textcolor{black}{with recursive usage of these multimodal encoders in our SEN pipeline.}

\subsection{Recursive Association (RA) of Super Neurons}

As mentioned in the introduction, we are partially inspired by the multimodal association procedure of the human brain~\cite{lansner2009associative,associativenature}.
When a human receives the input signal of one modality,
the brain area of this modality will produce neurotransmitters.
Then, these neurotransmitters are delivered to other brain areas that refer to other modalities.
Such a process is repeated many times to achieve multimodal association.
To mimic such a human-like process, we introduce a Recursive Association (RA) block for our SEN, \textcolor{black}{which serves as a lightweight, recursive feedback mechanism}.
As shown in Fig~\ref{fig:main structure}, it consists of three key components,
including knowledge integrating, distributing, and prompting of \textcolor{black}{multi-modal frozen encoders}.

\textbf{Knowledge Integrating}.
Suppose that,
we obtain the latent features $\{\mathbf{F}^{(i)}_{m}\}_{m=1}^{M}$ (i.e., knowledge) from the $i$-th \textcolor{black}{recursive step} of our SEN.
We first fuse all the latent features together as a multimodal knowledge $\mathbf{F}^{(i)}$:
\begin{equation}
\setlength\abovedisplayskip{1pt}
\setlength\belowdisplayskip{1pt}
\mathbf{F}^{(i)}=\mathcal{R}(\mathbf{F}^{(i)}_{1},...,\mathbf{F}^{(i)}_{m},...,\mathbf{F}^{(i)}_{M}).
\label{eq:integrate}
\end{equation}
All latent features are extracted from encoders of the foundation model.
Since these encoders are aligned via multimodal contrastive pretraining, the extracted features have the same dimension.
Hence, we use a fusion operation $\mathcal{R}$ (e.g., pooling or concatenation) to generate the multimodal knowledge.

\textbf{Knowledge Distributing}. After obtaining multimodal knowledge feature $\mathbf{F}^{(i)}$, we distribute it to all the super neurons in the \textcolor{black}{next recursive step},
in order to build up deeper multimodal interactions. 
Specifically, to facilitate deep interaction, we transform the aggregated knowledge $\mathbf{F}^{(i)}$ \textcolor{black}{into a specific context vector} for each individual modality format as Eq~\ref{eq:mntrans}. Where $\mathcal{H}^{(i)}_{m}$ is \textcolor{black}{a modality-specific MLP projector (validating the Sparse strategy)} to produce \textcolor{black}{the distributed context} of modality $m$ at the $i$-th \textcolor{black}{step}. 
\begin{equation}
\setlength\abovedisplayskip{1pt}
\setlength\belowdisplayskip{1pt}
\mathbf{G}^{(i)}_{m}=\mathcal{H}^{(i)}_{m}(\mathbf{F}^{(i)}).
\label{eq:mntrans}
\end{equation}

\textbf{Knowledge Prompting}.
The next question is how to \textcolor{black}{re-inject} the \textcolor{black}{distributed context} into the corresponding \textcolor{black}{foundation encoder} at the $i$-th \textcolor{black}{recursive step}.
Note that \textcolor{black}{these encoders are kept frozen to preserve the pre-trained knowledge manifold}.
Hence, each \textcolor{black}{encoder} has already had the input data of the corresponding modality $\mathbf{X}_{m}$.
In this case, we propose to convert the \textcolor{black}{context vector} into a \textcolor{black}{Dynamic Contextual Prompt} $\mathbf{P}^{(i)}_{m}$ for each modality,
\begin{equation}
\setlength\abovedisplayskip{1pt}
\setlength\belowdisplayskip{1pt}
\mathbf{P}^{(i)}_{m}=\mathbf{G}^{(i)}_{m} + \mathbf{Q}^{(i)}_{m}.
\label{eq:integrateP}
\end{equation}
For each modality at the $i$-th \textcolor{black}{step},
we introduce the learnable prompt $\mathbf{Q}^{(i)}_{m}$ to increase the adaptation capacity of our RA.
Finally, 
we concatenate the input data $\mathbf{X}_{m}$ and the contextual prompt $\mathbf{P}^{(i)}_{m}$, 
subsequently \textcolor{black}{re-injecting} them into the corresponding \textcolor{black}{encoder} to extract the latent feature in the next \textcolor{black}{step},
\begin{equation}
\setlength\abovedisplayskip{1pt}
\setlength\belowdisplayskip{1pt}
\mathbf{F}^{(i+1)}_{m}=\mathcal{M}_{m}(\mathbf{X}_{m} \oplus \mathbf{P}^{(i)}_{m}).
\label{eq:prompt}
\end{equation}
As we can see, the latent feature at the $(i+1)$-th \textcolor{black}{step} does not only depend on the input $\mathbf{X}_{m}$, but it also depends on $\mathbf{P}^{i}_{m}$ that is summarized from all the modalities at the $i$-th \textcolor{black}{step} (i.e., Eq.~\ref{eq:integrate}-\ref{eq:integrateP}).
In this case, this latent feature at the current \textcolor{black}{step} not only preserves semantic knowledge in this modality, but it also integrates multimodal contexts. 
\textcolor{black}{While standard foundation models effectively align encoders, they lack the mechanism for iterative refinement. By augmenting this alignment with recursive steps, SEN achieves a substantial Interaction Depth,} allowing the model to capture complex dynamics in open scenarios. 
To address this, we \textcolor{black}{perform recursive iterations}, allowing our SEN to achieve deep, multimodal interactions in videos.
In subsequent experiments, we demonstrate the necessity of this recursive process in enhancing the model's performance for downstream video tasks.

\subsection{SEN for Downstream Video Tasks}

To show the effectiveness of SEN, we next explain how to apply it for the four most representative video tasks,
including tracking, recognition, chatting, and editing.
Basically, we choose the simplest way for downstream task adaptation, where we add or concatenate the final feature of SEN into the corresponding task model.
This makes our SEN a plug-and-play unified encoder network for video understanding.
% Due to the limited space,
% we suggest readers to get more implementation details in the supp doc.

\begin{figure*}[t]
\centering
\vspace {-1.5em}
\includegraphics[width=\textwidth]{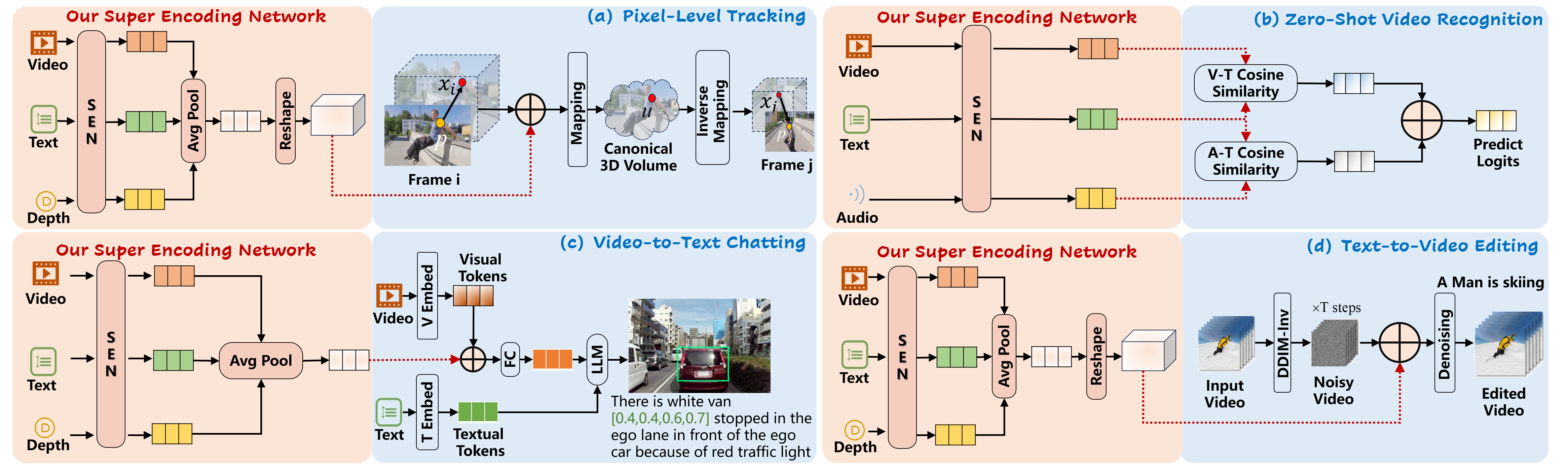}
\vspace {-1.5em}
\caption{\textcolor{black}{Illustration of SEN applied to downstream video tasks, from video tracking to video recognition, from video chatting to video editing.}}
\label{fig:ta}
\vspace {-1em}
\end{figure*}

\textbf{Pixel-Level Tracking}.
This task is to estimate full-length motion trajectories for every pixel in every frame of a video.
\textbf{First},
we choose the popular models~\cite{omnimotion,song2025track} for this task.
For convenience, we take Omnimotion~\cite{omnimotion} as an illustration.
As shown in Fig~\ref{fig:ta}(a), this model first specifies the location of a pixel $p_i$ at frame $i$, and concatenates it with a randomly initialized depth map to compose the 3D location of this pixel $x_i$.
Next, this model maps $x_i$ in frame $i$ to its canonical 3D volume, and leverages inverse mapping to find the corresponding $x_j$ in frame $j$ for pixel-level tracking. 
\textbf{Second}, besides video modality, we choose depth and text as other modalities in our SEN. The main reason is that depth can provide extra 3D information, and text can provide extra motion description of the target.
Both are important clues for this tracking task.
Hence, we use GPT-4o~\cite{gpt4} to generate the caption of this video, and use GLPN~\cite{kim2022global} to generate the depth of this video.
\textbf{Finally}, we feed all these modalities into our SEN for deeper interactions, and average the output features of all these modalities as a contextual feature.
Then, we resize it as the shape of $x_i$ in the Omnimotion model and add it to $x_i$ for tracking.
We freeze all the encoders in SEN and the task model and only fine-tune the learnable RA in SEN for task adaptation.

\textbf{Zero-Shot Video Recognition}.
This task is to identify the category of the query video from a number of given classes. To validate the generalization capacity of our SEN, we choose a zero-shot setting where the classes in the testing are not seen in the training data. 
\textbf{First}, the task model is the multimodal foundation model like ImageBind~\cite{imagebind}.
Hence, we can directly use multimodal encoders of the task model as super neurons of our SEN. 
\textbf{Second}, the input modalities of our SEN are video, text, and audio for the VggSound benchmark~\cite{vggsound}.
Zero-shot video recognition can be done by video-text and audio-text contrastive learning, i.e., we compute video-text and video-audio cosine similarity respectively, and add these two similarity matrices to get the overall similarity matrices for prediction as shown in Fig~\ref{fig:ta}(b).
\textbf{Finally}, we freeze all the encoders in SEN and the task model and only fine-tune the learnable RA in SEN for task adaptation.

% 这个任务名称之后更改
\noindent\textbf{Video-to-Text Chatting.}
\textcolor{black}{We select the Risk Object Localization and Intention and Suggestion Prediction (ROLISP) task~\cite{drama} to evaluate SEN on simultaneous video perception (localization) and reasoning (explanation).
\textbf{First}, we adopt the MLLM-based Shikra~\cite{shikra} as the task model, which comprises a vision encoder, an MLP connector, and an LLM (Fig.~\ref{fig:ta}(c)).
\textbf{Second}, to enhance spatial precision, we incorporate Depth (estimated via GLPN~\cite{kim2022global}) alongside Video and Text into SEN. The outputs from these modalities are averaged to form a unified contextual feature.
\textbf{Finally}, this feature is added to the visual tokens prior to the MLP connector. The enhanced tokens and user query are then processed by the LLM to generate the final bounding box and response. We fine-tune only the light-weighted RA module, keeping all other components frozen.}

\noindent\textbf{Text-to-Video Editing.}
\textcolor{black}{To assess the generalization capacity of SEN, we utilize the one-shot video editing setting where the model is trained on a single video-text pair.
\textbf{First}, we employ Tune-A-Video~\cite{tune-a-video} as the task model (Fig.~\ref{fig:ta}(d)). It is built upon Stable Diffusion and utilizes DDIM inversion~\cite{tune-a-video} to preserve the structural layout of the source video.
\textbf{Second}, we introduce Depth and Text (generated by GPT-4o~\cite{gpt4}) into SEN to provide explicit 3D structural cues and detailed semantic guidance.
\textbf{Finally}, the SEN output features are averaged and spatially resized to match the dimension of the noisy latent derived from DDIM inversion. This context is then added to the U-Net input for editing. Consistent with previous tasks, we freeze all encoders and the task model, optimizing only the learnable RA parameters.}

\section{Experiments}

\textbf{Datasets.}
For the Pixel-level tracking task, we utilized the DAVIS~\cite{davis} for evaluation as~\cite{omnimotion}.
For the video-level recognition task, we use the VGGSound~\cite{vggsound} dataset.
The DRAMA~\cite{drama} dataset is used to validate the video-to-text chatting task.
In the text-to-video editing task, we used 42 representative videos from the DAVIS~\cite{davis} dataset following~\cite{tune-a-video}.

\textcolor{black}{
\textbf{Metrics.} 
For \textbf{Pixel-level Tracking}, following~\cite{omnimotion}, we report Average Jaccard (AJ), Occlusion Accuracy (OA), and Positional Accuracy ($<\delta^x_\textrm{avg}$). 
We also measure Temporal Coherence (TC) via the $L_2$ distance of track accelerations (lower is better).
For \textbf{Zero-shot Recognition}, we report Top-1 Accuracy, Top-5 Accuracy, and Mean Average Precision (mAP).
\textbf{Video-to-Text Chatting} is evaluated using standard captioning metrics (BLEU-4, METEOR, CIDER, SPICE) and Mean Intersection Over Union (mIoU) for localization~\cite{hilm-d}.
For \textbf{One-shot Editing}, we measure Frame Consistency (average CLIP~\cite{clip} similarity between frame pairs) and Textual Alignment (CLIP score between output frames and the target prompt).}

\begin{table}[t]
    \centering

    \vspace{-1em}
    \resizebox{\linewidth}{!}{%
    \begin{tabular}{l l c c c}
        \toprule
        \textbf{Downstream Task} & \textbf{Sampling Strategy} & \textbf{Resolution} & \textbf{Training Time} & \textbf{Epochs/Steps} \\
        \midrule
        \textbf{Pixel-level Tracking} & Full video (30 FPS) & \multirow{2}{*}{$384 \times 512$} & $\sim$ 2.5 hours & \multirow{2}{*}{20k steps} \\
        \textit{(Target: Single Video Instance)} & (Dense Sampling) & & (per video) & \\
        \midrule
        \textbf{One-shot Video Editing} & Uniform (24 frames) & \multirow{2}{*}{$256 \times 256$} & $\sim$ 25 mins & \multirow{2}{*}{300 steps} \\
        \textit{(Target: Single Video Instance)} & (Sparse Sampling) & & (per video) & \\
        \midrule
        \textbf{Zero-shot Recognition} & Uniform (8 frames/clip) & \multirow{2}{*}{$224 \times 224$} & $\sim$ 1.5 hours & \multirow{2}{*}{2 epochs} \\
        \textit{(Dataset: VGGSound)} & (Clip-level) & & (total dataset) & \\
        \midrule
        \textbf{Video-to-Text Chatting} & Keyframe (2 FPS) & \multirow{2}{*}{$224 \times 224$} & $\sim$ 2.0 hours & \multirow{2}{*}{1 epochs} \\
        \textit{(Dataset: DRAMA)} & (Frame-level) & & (total dataset) & \\
        \bottomrule
    \end{tabular}
    }
    \vspace{-0.5em}
    \caption{\textcolor{black}{Implementation Details: Sampling, Resolution, and Training Cost.}}
    \label{tab:implementation_details}
    \vspace{-1.5em}
    
\end{table}

\textcolor{black}{
\textbf{Implementation Details.}
Across all tasks, we utilize consistent pre-processing to extract text and depth modalities. 
We employ \textbf{GPT-4o}~\cite{gpt4} (version 2024-05-13, temperature=1.0, top\_p=0.9) to generate dense semantic captions with 4 frames per video. 
For depth estimation, we use \textbf{GLPN-kitti} (62M)~\cite{kim2022global}, resizing inputs to $480 \times 640$ before feeding them into the ImageBind depth encoder. We report the sampling strategy, resolution, training time and training epoch/steps in Table~\ref{tab:implementation_details}. 
We employ \textbf{ImageBind} as the primary backbone for all tasks, leveraging its holistic embedding space that effectively aligns diverse modalities, including depth and audio. 
For the audio-centric VGGSound benchmark, we additionally utilize \textbf{VALOR} to benefit from its specialized audio-visual-text pre-training and robust Audio Spectrogram Transformer (AST) architecture. 
This strategic selection allows SEN to extract the most discriminative features, ranging from geometric structures to fine-grained acoustics, tailored to the specific requirements of each downstream scenario.
}

\textcolor{black}{
\textbf{Pixel-level Tracking.} 
We employ \textbf{ImageBind (1.1B)}~\cite{imagebind} as the frozen foundation model, utilizing its ViT-H (632M) Vision, OpenCLIP (354M) Text, and ViT-B (88M) Depth encoders. 
\textbf{Omnimotion}~\cite{omnimotion} serves as the task model. 
Following Omnimotion~\cite{omnimotion}, we use the AdamW optimizer with a cosine decay scheduler. The base learning rate is set to 3e-5, and the total training steps are 250k. The frame sampling interval gradually increases from 20 during training.}

\textcolor{black}{
\textbf{Zero-shot Recognition.} 
We evaluate SEN on both \textbf{VALOR} and \textbf{ImageBind} backbones. 
VALOR(0.5B) integrates CLIP-L~\cite{clip} (Vision), BERT-Base~\cite{bert} (Text), and AST~\cite{gong21b_interspeech} (Audio). 
We train for 10 epochs with a batch size of 32 and a learning rate of 1e-5. Images are resized to $224 \times 224$. The VALOR and ImageBind is also the task models.}

\textcolor{black}{
\textbf{Video-to-Text Chatting.} 
We integrate SEN (ImageBind backbone) with \textbf{Shikra}~\cite{shikra}. 
Following Shikra~\cite{shikra}, we utilize a cosine annealing scheduler with an initial learning rate of 2e-5 and a batch size of 16.}

\textcolor{black}{
\textbf{One-Shot Video Editing.} 
We adopt \textbf{Tune-A-Video}~\cite{tune-a-video} as the task model with the ImageBind-based SEN. 
The model is fine-tuned for 300 iterations with a learning rate of 3e-5 and a batch size of 1. 
Inference uses the DDIM sampler~\cite{ddim} with classifier-free guidance~\cite{cfdg}. 
All experiments were conducted on 8 NVIDIA RTX A6000 GPUs.}

\subsection{SOTA Comparison}

\begin{table*}[t]
    \begin{minipage}[t]{0.48\textwidth}
        \centering
        \small
        \setlength{\tabcolsep}{1.2mm}
        % 调整表格宽度适应页面
        \resizebox{\textwidth}{!}{
        \begin{tabular}{lcccc}
        \toprule
        Model & AJ~$\uparrow$ & $<\delta^x_\textrm{avg}$~$\uparrow$ & OA~$\uparrow$ & TC~$\downarrow$ \\ \midrule
        RAFT-D~\cite{raft} & 34.1 & 48.9 & 76.1 & 9.83 \\ 
        PIPs~\cite{pips} & 39.9 & 56.0 & 81.3 & 1.78 \\ 
        Flow-Walk-C~\cite{flow-walk} & 35.2 & 51.4 & 80.6 & 0.90 \\ 
        Deformable-Sprites~\cite{deformable-sprits} & 20.6 & 32.9 & 69.7 & 2.07 \\ 
        Omnimotion~\cite{omnimotion} & 51.7 & 67.5 & 85.3 & 0.74 \\ 
        CaDeX++~\cite{song2025track} & 59.4 & 77.4 & 85.9 & 0.68 \\ 
        \midrule
        \rowcolor{gray!20}
        \textbf{SEN + Omnimotion} & \textbf{54.4}\scriptsize{$\pm0.2$} & \textbf{68.8}\scriptsize{$\pm0.1$} & \textbf{86.6}\scriptsize{$\pm0.2$} & \textbf{0.70}\scriptsize{$\pm0.01$} \\ 
        \rowcolor{gray!20}
        \textbf{SEN + CaDeX++} & \textbf{62.0}\scriptsize{$\pm0.1$} & \textbf{78.0}\scriptsize{$\pm0.2$} & \textbf{87.8}\scriptsize{$\pm0.2$} & \textbf{0.62}\scriptsize{$\pm0.01$} \\ 
        \bottomrule
        \end{tabular}
        }
        \vspace{-0.6em}
        \caption{\textcolor{black}{SOTA Comparison on Pixel-level Tracking Task.}}
        \label{tab1:davis}
    \end{minipage}\hspace{1em}
    \begin{minipage}[t]{0.48\textwidth}
        \centering
        \setlength{\tabcolsep}{1mm}
        \small
        \renewcommand{\arraystretch}{0.9} %稍微调整行高以适应内容
        \resizebox{\textwidth}{!}{
        \begin{tabular}{lcccc}
        \toprule
        Method & Backbone & Top-1 Acc$\uparrow$ & Top-5 Acc$\uparrow$ & mAP$\uparrow$ \\ \midrule
        CLIP-Adapter~\cite{clipadapter} & VALOR & 46.31 & 72.45 & 54.20 \\ 
        CoCoop~\cite{Cocoop} & VALOR & 47.24 & 73.10 & 55.15 \\ 
        CoPL~\cite{goswami2024copl} & VALOR & 48.57 & 74.80 & 56.30 \\ 
        SEN & VALOR & 49.49\scriptsize{$\pm0.2$} & 76.92\scriptsize{$\pm0.2$} & 57.15\scriptsize{$\pm0.3$} \\ \midrule
        CLIP-Adapter~\cite{clipadapter} & ImageBind & 50.93 & 78.50 & 58.20 \\ 
        CoCoop~\cite{Cocoop} & ImageBind & 51.40 & 79.25 & 59.10 \\ 
        CoPL~\cite{goswami2024copl} & ImageBind & 51.67 & 79.80 & 59.40 \\ 
        SEN & ImageBind & 53.36\scriptsize{$\pm0.2$} & 83.42\scriptsize{$\pm0.2$} & 61.80\scriptsize{$\pm0.3$} \\ \midrule
        \rowcolor{gray!20}
        \textbf{SEN} & \textbf{\begin{tabular}[c]{@{}c@{}}ImageBind\\ +VALOR  \end{tabular}} & \textbf{53.79}\scriptsize{$\pm0.19$} & \textbf{84.15}\scriptsize{$\pm0.14$} & \textbf{62.30}\scriptsize{$\pm0.18$} \\ \bottomrule
        \end{tabular}
        }
        \vspace{-0.6em}
        \caption{\textcolor{black}{SOTA Comparison on VGGSound Zero-Shot Recognition.}}
        \label{tab2:vggsound}
    \end{minipage}
    \vspace{-1em}
\end{table*}

\begin{table*}[!]
	\centering
	\begin{minipage}[!]{0.49\textwidth}
	\small
        \centering
        \setlength{\tabcolsep}{1mm}

        \resizebox{\textwidth}{!}{
        \begin{tabular}{cccccc} \toprule
        Model & B$\uparrow$ & M$\uparrow$ & C$\uparrow$ & S$\uparrow$ & mIoU$\uparrow$ \\ \midrule
        Video-Chat~\cite{videochat} & 36.8 & 29.0 & 153.0 & 43.5 & --- \\ 
        Video-LLaMA~\cite{video-llama} & 46.0 & 38.4 & 253.5 & 38.7 & 42.8 \\ 
        BLIP-2~\cite{blip2} & 50.6 & 38.1 & 258.5 & 53.7 & 46.3 \\ 
        Video-LLaMA2~\cite{cheng2024videollama2} &  51.3 & 38.5 & 305.9 & 54.2 & 63.8 \\ 
        Shikra~\cite{shikra} & 51.6 & 38.4 & 310.1 & 54.5 & 64.2 \\ 
        \midrule
        \rowcolor{gray!20}
        \textbf{SEN + Shikra} & \textbf{52.9}\scriptsize{$\pm$0.2} & \textbf{39.0}\scriptsize{$\pm$0.1} & \textbf{325.5}\scriptsize{$\pm$0.3} & \textbf{55.7}\scriptsize{$\pm$0.2} & \textbf{66.9}\scriptsize{$\pm$0.2} \\  \bottomrule
        \end{tabular}
        }
        \vspace{-0.6em} 
        \caption{\textcolor{black}{SOTA Comparison on DRAMA Video-to-Text Chatting on Automatic Driving Scenario.}}
        \vspace{-1em}

        \label{tab3: drama}
	\end{minipage}\hspace{1em}
	\begin{minipage}[!]{0.47\textwidth}
	\small\centering
    \setlength{\tabcolsep}{1mm}
        \renewcommand{\arraystretch}{0.8}
        \setlength{\tabcolsep}{1mm}
        
        \resizebox{\textwidth}{!}{
        \begin{tabular}{ccc} \toprule
        Method  &  \begin{tabular}[c]{@{}c@{}}Frame\\ Consistency \end{tabular} $\uparrow$ & \begin{tabular}[c]{@{}c@{}}Textual\\ alignment\end{tabular}  $\uparrow$  \\ \midrule
        CogVideo~\cite{cogvideo} & 90.64 & 23.91 \\ 
        Plug-and-Play~\cite{plug-and-play} & 88.89 & 27.56 \\ 
        AnyV2V~\cite{ku2024anyv2v} & 92.26 & 26.51   \\
        Tune-A-Video~\cite{tune-a-video} & 92.40 & 27.58 \\ 
        UniEdit~\cite{bai2024uniedit}& 93.08 & 27.96  \\
        \midrule
        \rowcolor{gray!20}
        \textbf{SEN + Tune-A-Video} & \textbf{96.53}\scriptsize{$\pm$0.15} & \textbf{34.01}\scriptsize{$\pm$0.25} \\ \bottomrule
        \end{tabular} }
        \vspace{-0.6em} 
        \caption{\textcolor{black}{SOTA Comparison on One-shot Video Editing Task.}}
        \vspace{-1em}
        \label{tab4:clip score}
 	\end{minipage}
\end{table*}

\textcolor{black}{
\textbf{Pixel-level Tracking Task.} 
As shown in Table~\ref{tab1:davis}, SEN achieves state-of-the-art performance compared to models without domain-specific pretraining. 
Integrating SEN with Omnimotion~\cite{omnimotion} and CaDeX++~\cite{song2025track} significantly boosts their tracking accuracy. 
Specifically, on CaDeX++, SEN improves the Average Jaccard (AJ) by 2.6\% and reduces Temporal Coherence (TC) by 8.8\% (0.68 $\rightarrow$ 0.62), demonstrating enhanced temporal consistency.}

\textcolor{black}{\textbf{Video-level Zero-shot Recognition.} 
We compared SEN against parameter-efficient fine-tuning methods (CLIP-Adapter~\cite{clipadapter}, CoPL~\cite{goswami2024copl}, CocoOp~\cite{Cocoop}) on VGGSound. 
For a fair comparison, all baselines were extended with the AST~\cite{gong21b_interspeech} audio encoder. 
Table~\ref{tab2:vggsound} shows that SEN outperforms these approaches across both VALOR and ImageBind backbones. 
Furthermore, scaling SEN by combining encoders from both foundation models yields additional performance gains.}

\begin{table*}[t]
    \vspace{-1em}
    \centering
    \small
    \setlength{\tabcolsep}{0.5mm}
     \resizebox{\textwidth}{!}{
    % 修改这里：lc 后面原本有11个c，现在增加到12个c，总共14列
   \begin{tabular}{lc cccccccccccc} \toprule
         \multirow{3}{*}{\begin{tabular}[c]{@{}c@{}}Fusion\\ Format\end{tabular}} & \multirow{3}{*}{Params}  &  \multicolumn{4}{c}{{DAVIS Pixel-level Tracking}}  & VGGSound & \multicolumn{5}{c}{{Drama Automatic Driving Chatting}} & \multicolumn{2}{c}{{DAVIS One-shot Video Editing}}  \\ \cmidrule{3-14}
         & & AJ$\uparrow$ & $<\delta^x_\textrm{avg}$$\uparrow$ & OA$\uparrow$ & TC$\downarrow$ & Top1 Acc$\uparrow$ & B$\uparrow$ & M$\uparrow$ & C$\uparrow$ & S$\uparrow$ & mIoU$\uparrow$ &  \begin{tabular}[c]{@{}c@{}}Frame\\ Consistency$\uparrow$\end{tabular} & \begin{tabular}[c]{@{}c@{}}Textual\\ alignment$\uparrow$\end{tabular}   \\ \midrule
         Task Model & - & 51.7 & 67.5 & 85.3 & 0.74 & 51.67 & 51.6 & 38.4 & 310.1 & 54.5 & 64.2 & 92.40 & 27.58     \\ 
        Baseline & $\sim$0.6M & 51.9 & 67.7 & 85.6 & 0.74 & 51.96 & 51.8 & 38.5 & 312.4 & 54.7 & 64.3 & 92.95 & 28.48     \\ 
        Transformer~\cite{vit} & $\sim$7.1M & 52.2 & 67.8 & 85.5 & 0.73 & 51.25 & 52.5 & 38.7 & 320.2 & 55.5 & 65.8 & 94.96 & 33.02     \\ 
       \begin{tabular}[c]{@{}c@{}}Prompt-based\\ Adapter~\cite{vpt}\end{tabular}  & $\sim$0.1M & 51.8 & 67.6 & 85.4 & 0.74 & 51.92 & 51.7 & 38.5 & 313.5 & 54.8 & 64.5 & 93.50 & 29.10     \\ 
        Perceiver Style~\cite{perceiver} & $\sim$4.5M & 52.1 & 67.9 & 85.5 & 0.73 & 52.45 & 52.0 & 38.6 & 318.5 & 55.0 & 65.2 & 94.20 & 31.80     \\ 
        Adapter~\cite{clipadapter} & $\sim$1.2M & 52.6 & 68.0 & 85.8 & 0.72 & 52.51 & 52.2 & 38.7 & 320.5 & 55.2 & 65.6 & 94.80 & 32.85  \\ 
        \rowcolor{gray!20}
        \textbf{RA (Ours)} & \textbf{$\sim$1.75M} & \textbf{54.4} & \textbf{68.8} & \textbf{86.6} & \textbf{0.70} & \textbf{53.36} & \textbf{52.9} & \textbf{39.0} & \textbf{325.5} & \textbf{55.7} & \textbf{66.9} & \textbf{96.53} & \textbf{34.01} \\ \bottomrule
    \end{tabular}
    }
    \vspace{-0.6em} 
    \caption{\textcolor{black}{Comparison of different feature fusion formats. \textbf{Baseline} refers to simple concatenation. Our \textbf{RA} significantly outperforms other mechanisms.}}
    \vspace{-0.5em}
    \label{layer_format}
\end{table*}

\textcolor{black}{\textbf{Video-to-Text Chatting on Automatic Driving Scenario.} 
Table~\ref{tab3: drama} highlights the superior spatial-semantic understanding of SEN. 
It achieves a 5.0\% improvement in CIDER score (310.1 $\rightarrow$ 325.5) for captioning and a 2.7\% increase in mIoU for risk object localization, validating the benefit of deeper multimodal interaction.}

\textcolor{black}{\textbf{Text-to-Video One-shot Editing.} 
Following the protocol of Tune-A-Video~\cite{tune-a-video}, Table~\ref{tab4:clip score} reports that SEN improves Textual Alignment by 4.5\% and Frame Consistency by 0.8\%. 
These results confirm that our recursive association effectively preserves temporal structure while enhancing semantic alignment in video generation.}

\subsection{Ablations}

% TODO:这里需要增加一些perceiver等表格上的细节

\textbf{Different Feature Fusion Format.}
\textcolor{black}{
We conduct a comprehensive ablation in Table~\ref{layer_format} to isolate the efficacy of our Recursive Association (RA) block. 
First, we establish a \textbf{Fair Baseline} utilizing the \textbf{exact same multimodal} }
\begin{table*}[t]
    \centering
    \small
    % \vspace{-1em}
    \setlength{\tabcolsep}{0.5mm}
    \resizebox{\textwidth}{!}{%
    \begin{tabular}{lcccccccccccc} \toprule
         \multirow{3}{*}{\begin{tabular}[c]{@{}c@{}}Caption\\ Model\end{tabular}}  &  \multicolumn{4}{c}{{DAVIS Pixel-level Tracking}}  & VGGSound & \multicolumn{5}{c}{{Drama Automatic Driving Chatting}} & \multicolumn{2}{c}{{DAVIS One-shot Video Editing}}  \\ \cmidrule(lr){2-13}
         & AJ$\uparrow$ & $<\delta^x_\textrm{avg}$$\uparrow$ & OA$\uparrow$ & TC$\downarrow$ & Top1 Acc$\uparrow$ & B$\uparrow$ & M$\uparrow$ & C$\uparrow$ & S$\uparrow$ & mIoU$\uparrow$ &  \begin{tabular}[c]{@{}c@{}}Frame\\ Consistency$\uparrow$\end{tabular} & \begin{tabular}[c]{@{}c@{}}Textual\\ alignment$\uparrow$\end{tabular}   \\ \midrule
        Task Model  & 51.7 & 67.5 & 85.3 & 0.74 & 51.67 & 51.6 & 38.4 & 310.1 & 54.5 & 64.2 & 92.40 & 27.58     \\ 
        Qwen2.5-VL-72B~\cite{qwen2.5vl} & 54.3 & 68.7 & 86.5 & 0.70 & 53.41 & 52.9 & 38.9 & 324.8 & 55.7 & 66.9 & 96.50 & 33.94 \\
        GPT-4o~\cite{gpt4} & 54.4 & 68.8 & 86.6 & 0.70 & 53.36 & 52.9 & 39.0 & 325.5 & 55.7 & 66.9 & 96.53 & 34.01 \\ 
        \bottomrule
    \end{tabular}
    }
    \vspace{-0.5em}
    \caption{\textcolor{black}{Ablation for different caption models.}}
    \label{tab:caption_ablation}
    \vspace{-1em}
\end{table*}
\begin{wraptable}{l}{0.45\linewidth}
    \centering
    \small
    \setlength{\tabcolsep}{1mm}
    \vspace{-1em}
    
    \resizebox{\linewidth}{!}{%
    \begin{tabular}{l c c}
        \toprule
        \textbf{Method}  & \textbf{Training Source} & \textbf{Top-1 Acc} \\
        \midrule
    ImageBind~\cite{imagebind}  & ImageBind Pretrained & 78.9\% \\
        \rowcolor{gray!15} \textbf{SEN (Ours)} & \textbf{VGGSound Transfer} & \textbf{82.1\%} \\
        \bottomrule
    \end{tabular}
    }
    \vspace{-0.8em}
     \caption{\textcolor{black}{Zero-Shot Transfer Performance on UCF101.}}
    \label{tab:ucf101_zeroshot}
    \vspace{-2em}
    
\end{wraptable}
\textcolor{black}{\textbf{inputs as SEN} but via simple concatenation and MLP projection. 
The significant performance gap confirms that our gains derive from the recursive interaction design rather than merely additional data.
We further benchmark RA against four representative fusion paradigms: 
(1) \textbf{Transformer}~\cite{vit}, which employs Multi-Head Cross-Attention to aggregate features;
(2) \textbf{Prompt-based Adaptation}~\cite{vpt}, prepending learnable continuous tokens to encoders; 
(3) \textbf{Standard Adapters}~\cite{clipadapter}, inserting bottleneck (Down-ReLU-Up) layers; 
and (4) \textbf{Perceiver-style Hub}~\cite{perceiver}, compressing inputs into fixed latents (64 tokens).
Results indicate that Perceiver-style hubs lose fine-grained spatial cues due to compression, while heavy Transformers ($\sim$7.1M params) suffer from overfitting in data-scarce tasks. 
In contrast, our RA ($\sim$1.75M params) outperforms these static or parameter-heavy alternatives, achieving an optimal balance between deep recursive interaction and parameter efficiency.}

% TODO: 要修改这一段，让整体看起来更好
\textbf{Different Caption Models.}
\textcolor{black}{
To verify the robustness of SEN regarding the choice of captioning models and ensure reproducibility, we conducted an ablation study using the open-source \textbf{Qwen2.5-VL-72B}~\cite{qwen2.5vl} as a substitute for the proprietary GPT-4o~\cite{gpt4}, with all other configurations remaining identical. 
As presented in \textbf{Table~\ref{tab:caption_ablation}}, the performance gap between using the open-source and proprietary models is marginal, thereby eliminating concerns regarding reliance on commercial APIs. 
Crucially, even with the open-source captioner, SEN maintains a significant performance advantage over the Task Model. 
This conclusively proves that our improvements are primarily driven by the efficacy of the \textbf{Recursive Association (RA) architecture}, rather than the specific quality of the input captions.}

\begin{table*}[t]
    \renewcommand{\arraystretch}{0.9}
    \centering
    \small
    % \vspace{-1em}  % 如需调整表格与正文间距可取消注释
    
    \resizebox{\textwidth}{!}{
    \begin{tabular}{ccccccccccccc} \toprule
         \multirow{3}{*}{\begin{tabular}[c]{@{}c@{}}Num\\ RA\end{tabular}}   &  \multicolumn{4}{c}{{DAVIS Pixel-level Tracking}}  & VGGSound & \multicolumn{5}{c}{{Drama Automatic Driving Chatting}} & \multicolumn{2}{c}{{DAVIS One-shot Video Editing}}  \\ \cmidrule{2-13}
         &  AJ$\uparrow$ & $<\delta^x_\textrm{avg}$$\uparrow$ & OA$\uparrow$ & TC$\downarrow$ & Top1 Acc$\uparrow$ & B$\uparrow$ & M$\uparrow$ & C$\uparrow$ & S$\uparrow$ & mIoU$\uparrow$ &  \begin{tabular}[c]{@{}c@{}}Frame\\ Consistency$\uparrow$\end{tabular} & \begin{tabular}[c]{@{}c@{}}Textual\\ alignment$\uparrow$\end{tabular}   \\ \midrule
        1  & 52.0 & 67.7 & 85.4 & 0.74 & 50.93 & 52.4 & 38.6 & 319.4 & 55.4 & 65.5 & 94.85 & 32.48   \\ 
        2  & 53.0 & 68.3 & 85.6 & 0.71 & 52.51 & 52.9 & 38.6 & 322.2 & 55.5 & 66.4 & 95.01 & 32.51   \\ 
        \rowcolor{gray!20}
        \textbf{3}   & \textbf{54.4} & \textbf{68.8} & \textbf{86.6} & \textbf{0.70} & \textbf{53.36} & \textbf{52.9} & \textbf{39.0} & \textbf{325.5} & \textbf{55.7} & \textbf{66.9} & \textbf{96.53} & \textbf{34.01} \\ 
        4  & 54.4 & 68.7 & 86.6 & 0.71 & 53.39 & 52.8 & 38.9 & 325.6 & 55.6 & 66.4 & 96.60 & 33.98   \\ \bottomrule
    \end{tabular}}
    \vspace{-0.6em} 
    
    \caption{Necessity of Recursive Design and Upper Bound.}
    \vspace{-1em} 
    \label{tab:num_layer}
\end{table*}

\textbf{Zero-shot Transfer Performance on UCF101.}
\textcolor{black}{To validate the robustness of our learned interactions, we conduct a cross-dataset zero-shot transfer experiment. We directly evaluate the SEN model trained on VGGSound (Audio-Visual Recognition) on the UCF101 dataset~\cite{soomro2012ucf101} (Action Recognition) without any retraining. As shown in Table~\ref{tab:ucf101_zeroshot}, simply applying SEN yields a \textbf{3.2\%} improvement over the strong ImageBind baseline. This result highlights that our RA blocks capture generic multimodal interaction patterns rather than merely overfitting to the training distribution, enabling effective transfer across different video domains.}

\textbf{Necessity of Recursive Design and Upper Bound.} 
\textcolor{black}{
To validate the necessity of our recursive design, we analyze the impact of the number of RA blocks (recursive steps $K$) as shown in Table~\ref{tab:num_layer}. 
Each RA block represents a round of multimodal interaction. 
The results demonstrate a clear trend: deepening the multimodal interaction enables the model to better explore cross-modal contexts, thereby enhancing robustness and task performance.
\textit{We further investigate the saturation point of this recursive depth.} 
As shown in Table~\ref{tab:num_layer}, performance tends to converge when the number of RA blocks reaches $K=3$. 
This can be explained by the \textbf{effective depth} of our architecture. 
Since each recursive step in SEN re-utilizes the entire foundation encoder (e.g., the 32-layer ViT-H/14~\cite{vit,clip}), a setting of 3 RA Blocks is mathematically equivalent to traversing a network depth of $3 \times 32 = 96$ attention blocks.
This depth is sufficient to encode deep multimodal interactions. 
Increasing $K$ further ($K>3$) yields diminishing returns and introduces risks of overfitting and increased inference latency. 
Therefore, we adopt 3 RA blocks as the optimal configuration.}

\begin{table*}[t]
    \renewcommand{\arraystretch}{1}
    \centering
    \small
    % \caption{Prompt preprocess setting.}
    % \vspace{-2em}
    \resizebox{\textwidth}{!}{
    \begin{tabular}{ccccccccccccc} \toprule
          \multirow{3}{*}{Method}   &  \multicolumn{4}{c}{{DAVIS Pixel-level Tracking}}  & VGGSound & \multicolumn{5}{c}{{Drama Automatic Driving Chatting}} & \multicolumn{2}{c}{{DAVIS One-shot Video Editing}}  \\ \cmidrule{2-13}
         &  $AJ\uparrow$ & $<\delta^x_{\textrm{avg}}\uparrow$ & $OA\uparrow$ & $TC\downarrow$ & $Top1\ Acc\uparrow$ & $B\uparrow$ & $M\uparrow$ & $C\uparrow$ & $S\uparrow$ & $mIoU\uparrow$ &  \begin{tabular}[c]{@{}c@{}}Frame\\ Consistency \end{tabular} $\uparrow$ & \begin{tabular}[c]{@{}c@{}}Textual\\ alignment\end{tabular}  $\uparrow$   \\ \midrule
        Add & 51.5 & 67.5 & 85.4 & 0.70 & 52.70 & 52.4 & 38.6 & 323.0 & 55.0 & 66.6 & 95.77 & 33.14 \\ 
        Attention & 51.3 & 65.7 & 86.2 & 0.71 & 51.97  & 52.3 & 38.5 & 320.7 & 55.3 & 66.6 & 95.60 & 33.25 \\ 
        MoE & 52.4 & 68.2 & 85.0 & 0.72 & 52.55  & 52.5 & 38.6 & 320.1 & 55.0 & 66.4 & 95.89 & 33.10 \\
        Concat  & 54.2 & 68.7 & 86.4 & 0.71 & 52.98  & 52.7 & 38.9 & 323.6 & 55.5 & 66.7 & 95.36 & 33.82 \\ 
        \rowcolor{gray!20}  \textbf{Avg}  & \textbf{54.4} & \textbf{68.8} & \textbf{86.6} & \textbf{0.70} & \textbf{53.36}  & \textbf{52.9} & \textbf{39.0} & \textbf{325.5} & \textbf{55.7} & \textbf{66.9} & \textbf{96.53} & \textbf{34.01} \\ 
        \bottomrule
    \end{tabular}
    }
    \vspace{-0.6em} 
    
    \caption{Knowledge Integrating.}
    \vspace{-1.2em}
    
    \label{prompt preprocess}
    
\end{table*}

% TODO:增加这部分的实施细节。
\textbf{Knowledge Integrating.} 
\textcolor{black}{
We investigate the optimal strategy for the Knowledge Integrating component ($\mathcal{R}$) by comparing five distinct mechanisms in Table~\ref{prompt preprocess}. 
Besides standard \textbf{Element-wise Summation (Add)} and \textbf{Concatenation (Concat)}, we evaluate: 
(1) \textbf{Attention}, which utilizes a Transformer layer to model cross-modal dependencies; 
(2) \textbf{MoE (Mixture of Experts)}, which employs a lightweight router to predict dynamic weights; 
and (3) \textbf{Average Pooling (Avg)}. 
Results indicate that parameter-heavy methods (Attention, MoE) do not outperform simpler ones. 
While Concat yields competitive results, \textbf{Avg} consistently demonstrates superior robustness across all tasks without introducing additional parameters. 
Consequently, we adopt Avg as our default integration method for its balance of efficiency and performance.}

\begin{table*}[t]
    \centering
    \small
    \resizebox{\textwidth}{!}{
    \begin{tabular}{ccccccccccccc} \toprule
         \multirow{3}{*}{Method}   &  \multicolumn{4}{c}{{DAVIS Pixel - level Tracking}}  & VGGSound & \multicolumn{5}{c}{{Drama Automatic Driving Chatting}} & \multicolumn{2}{c}{{DAVIS One - shot Video Editing}}  \\ \cmidrule{2 - 13}
         &  $AJ\uparrow$ & $<\delta^x_{\textrm{avg}}\uparrow$ & $OA\uparrow$ & $TC\downarrow$ & $Top1\ Acc\uparrow$ & $B\uparrow$ & $M\uparrow$ & $C\uparrow$ & $S\uparrow$ & $mIoU\uparrow$ &  \begin{tabular}[c]{@{}c@{}}Frame\\ Consistency \end{tabular} $\uparrow$ & \begin{tabular}[c]{@{}c@{}}Textual\\ alignment\end{tabular}  $\uparrow$   \\ \midrule
        Dense & 52.0 & 68.1 & 86.5 & 0.71 & 52.68  & 52.9 & 38.9 & 325.4 & 55.7 & 66.8 & 96.49 & 33.94\\  
        % MoE & ~ & ~ & ~ & ~ & ~ & ~ & ~ & ~ & ~ & ~ & ~ &  \\ 
        % Medium & ~ & ~ & ~ & ~ & ~ & ~ & ~ & ~ & ~ & ~ & ~ &  \\ 
        \rowcolor{gray!20} \textbf{Sparse} & \textbf{54.4} & \textbf{68.8} & \textbf{86.6} & \textbf{0.70} & \textbf{53.36}  & \textbf{52.9} & \textbf{39.0} & \textbf{325.5} & \textbf{55.7} & \textbf{66.9} & \textbf{96.53} & \textbf{34.01}  \\ 
        \bottomrule
    \end{tabular}}
    %\vspace {-0.6em}
    \vspace{-0.6em} 
    \caption{Knowledge Distributing.}
    \vspace {-1.5em}
    \label{prompt dis}
    
\end{table*}

\textbf{Knowledge Distributing.} 
\textcolor{black}{Table~\ref{prompt dis} compares two distribution strategies: \textit{Sparse} and \textit{Dense}. 
Specifically, the \textit{Sparse} method employs a specialized MLP $\mathcal{H}^{(i)}_{m}$ for each modality as shown in Fig~\ref{fig:main structure}, whereas the \textit{Dense} method utilizes a single unified MLP $\mathcal{H}$ for all features.
Sparse method gets the better results. 
For instance, it outperforms the Dense baseline by 2.4\% in AJ on DAVIS tracking (54.4 vs. 52.0). 
This demonstrates that using modality-specific projections is more effective for distributing multimodal context than a shared mapping.}

\textbf{Knowledge Prompting.} 
\textcolor{black}{
Table~\ref{tab:learnable} investigates the efficacy of the learnable prompt $\mathbf{Q}^{i}_{m}$. 
The results show that incorporating $\mathbf{Q}^{i}_{m}$ consistently yields performance gains, such as a 1.0 point improvement in AJ on DAVIS (53.4 $\rightarrow$ 54.4) and a 0.2\% rise in Top-1 Acc on VGGSound. 
This validates that learnable prompts effectively enhance the adaptation capacity of our RA module, aligning with the principles proposed in~\cite{CoOp}.}

\begin{table*}[t]
    \renewcommand{\arraystretch}{1}
    \centering
    \small
    %\vspace{-0.3em}
    % \vspace {-2em}
    \resizebox{\textwidth}{!}{
    \begin{tabular}{ccccccccccccc} \toprule
          \multirow{3}{*}{\begin{tabular}[c]{@{}c@{}}Prompting\\ Method\end{tabular}}   &  \multicolumn{4}{c}{{DAVIS Pixel-level Tracking}}  & VGGSound & \multicolumn{5}{c}{{Drama Automatic Driving Chatting}} & \multicolumn{2}{c}{{DAVIS One-shot Video Editing}}  \\ \cmidrule{2-13}
         &  $AJ\uparrow$ & $<\delta^x_{\textrm{avg}}\uparrow$ & $OA\uparrow$ & $TC\downarrow$ & $Top1\ Acc\uparrow$ & $B\uparrow$ & $M\uparrow$ & $C\uparrow$ & $S\uparrow$ & $mIoU\uparrow$ &  \begin{tabular}[c]{@{}c@{}}Frame\\ Consistency \end{tabular} $\uparrow$ & \begin{tabular}[c]{@{}c@{}}Textual\\ alignment\end{tabular}  $\uparrow$   \\ \midrule
         \rowcolor{gray!20}
        w/ $\mathbf{Q}^{i}_{m}$  & \textbf{54.4} & \textbf{68.8} & \textbf{86.6} & \textbf{0.70}   & \textbf{53.36} & \textbf{52.9} & \textbf{39.0} & \textbf{325.5} & \textbf{55.7} & \textbf{66.9}  & \textbf{96.53} & \textbf{34.01} \\ 
        w/o $\mathbf{Q}^{i}_{m}$  & 53.4 & 68.2 & 86.0 & 0.71 & 53.16 & 52.7 & 38.8 & 323.5 & 55.5 & 66.3 & 95.81 & 33.45 \\ 
        \bottomrule
    \end{tabular}}
    \vspace {-0.6em}
    
    \caption{Knowledge Prompting.}
    \label{tab:learnable}
    \vspace {-0.6em}
\end{table*}
\begin{table*}[!]
    \centering
    \small
    \resizebox{\textwidth}{!}{
    \begin{tabular}{c|cccc|ccccc|cc} \toprule
         \multirow{3}{*}{\begin{tabular}[c]{@{}c@{}}Video\\ Encoder \end{tabular}}   &  \multicolumn{4}{c|}{{DAVIS Pixel-level Tracking}}   & \multicolumn{5}{c|}{{Drama Automatic Driving Chatting}} & \multicolumn{2}{c}{{DAVIS One-shot Video Editing}}  \\ \cmidrule{2-12}
         &  AJ$\uparrow$ & $<\delta^x_\textrm{avg}$$\uparrow$ & OA$\uparrow$ & TC$\downarrow$  & B$\uparrow$ & M$\uparrow$ & C$\uparrow$ & S$\uparrow$ & mIoU$\uparrow$ &  \begin{tabular}[c]{@{}c@{}}Frame\\ Consistency$\uparrow$\end{tabular} & \begin{tabular}[c]{@{}c@{}}Textual\\ alignment$\uparrow$\end{tabular}   \\ \midrule
         
         \rowcolor{gray!10} \textit{Task Model Size} & \multicolumn{4}{c|}{\textit{$\approx$ 8.7M}}  & \multicolumn{5}{c|}{\textit{$\approx$ 7B}} & \multicolumn{2}{c}{\textit{$\approx$ 1.1B}} \\
         \midrule
         \multicolumn{12}{l}{\textit{\textbf{1. Ablation on Video Encoder (Text encoder: OpenCLIP-H, Depth Encoder: Vit-B/16)}}} \\
        ViT-B/16(88M)~\cite{vit} & 51.2 & 65.6 & 84.8 & 0.72  & 52.1 & 38.4 & 319.8 & 54.8 & 66.3 & 95.62 & 33.21\\  
        ViT-L/14(304M)~\cite{vit} & 52.8 & 68.4 & 86.2 & 0.71   & 52.7 & 38.8 & 320.9 & 55.1 & 66.6 & 96.24 & 33.62\\  
        UniformerV2-L(354M)~\cite{uniformerv2}  & 52.9 & 68.5 & 86.4 & 0.70   & 52.8 & 38.9 & 321.2 & 55.2 & 66.8 & 96.30 & 33.68\\ 
        ViT-H/14(632M)~\cite{vit}$^{\dagger}$ & 54.4 & 68.8 & 86.6 & 0.70   & 52.9 & 39.0 & 325.5 & 55.7 & 66.9 & 96.53 & 34.01  \\ 
        \rowcolor{gray!20} 
        
        \textbf{InternVideo2(1B)}~\cite{internvideo2} & \textbf{54.6} & \textbf{68.9} & \textbf{86.8} & \textbf{0.69}   & \textbf{53.1} & \textbf{39.2} & \textbf{325.6} & \textbf{55.7} & \textbf{66.9} & \textbf{96.72} & \textbf{33.95}\\ 
        \bottomrule
        \multicolumn{12}{l}{\textit{\textbf{2. Ablation on Text Encoder (Video encoder: Vit-H/14, Text encoder: OpenCLIP-H/14}}} \\
        BERT-Base (109M) & 53.8 & 68.1 & 86.1 & 0.72 & 51.4 & 38.1 & 318.5 & 54.2 & 65.8 & 95.82 & 32.90 \\
        OpenCLIP-L/14 (123M) & 54.1 & 68.4 & 86.3 & 0.71 & 52.2 & 38.6 & 322.8 & 55.1 & 66.4 & 96.15 & 33.45 \\
        \rowcolor{gray!15} 
        \textbf{OpenCLIP-H/14 (354M)} $^{\dagger}$ & \textbf{54.4} & \textbf{68.8} & \textbf{86.6} & \textbf{0.70} & \textbf{52.9} & \textbf{39.0} & \textbf{325.5} & \textbf{55.7} & \textbf{66.9} & \textbf{96.53} & \textbf{34.01} \\
        \midrule
        \multicolumn{12}{l}{\textit{\textbf{3. Ablation on Depth Encoder (Video encoder: Vit-H/14, Depth Encoder: Vit-B/16)}}} \\
        ViT-S/16 (22M) & 53.2 & 67.9 & 85.9 & 0.73 & 52.5 & 38.7 & 323.1 & 55.3 & 66.1 & 96.10 & 33.78 \\
        \textbf{ViT-B/16 (88M)} $^{\dagger}$ & 54.4 & 68.8 & 86.6 & 0.70 & 52.9 & 39.0 & 325.5 & 55.7 & 66.9 & 96.53 & 34.01 \\
        \rowcolor{gray!20} 
        
        ViT-L/14 (304M) & \textbf{54.5} & \textbf{68.9} & \textbf{86.7} & \textbf{0.69} & \textbf{53.1} & \textbf{39.2} & \textbf{325.8} & \textbf{56.0} & \textbf{67.1} & \textbf{96.60} & \textbf{34.11} \\
        \bottomrule
    \end{tabular}}
    \caption{\textcolor{black}{Ablation Study for Different Modality Encoders of SEN. $^{\dagger}$ denotes the default setting in our main paper.}}
    \vspace{-1.2em}
    \label{tab:abl_video_encoder}
\end{table*}

\textbf{Ablation for Different Encoders.} 
\textcolor{black}{
Table~\ref{tab:abl_video_encoder} presents a comprehensive ablation study across Video, Text, and Depth encoders to validate the robustness of SEN.
\textbf{1) On Video Encoders:} We observe a positive correlation between performance and model capacity. 
Notably, \textbf{InternVideo2-1B}~\cite{internvideo2}, despite lacking pre-alignment with the frozen text/depth encoders, achieves state-of-the-art results. 
This confirms the generality of SEN: it effectively associates encoders from different sources without requiring strict pre-training alignment.
\textbf{2) On Text Encoders:} Employing a stronger encoder (text encoder in OpenCLIP-H~\cite{clip}) significantly boosts performance on semantic-heavy tasks such as Video Chatting.
\textbf{3) On Depth Encoders:} Increasing depth encoder capacity (to ViT-L) yields further gains in geometrically sensitive tasks like Pixel-level Tracking (AJ \textbf{53.2 $\to$ 54.5}), demonstrating the benefit of finer spatial structural cues.
Overall, stronger encoders consistently enhance performance across diverse task model scales.}

\begin{table}[t]
    \centering
    \begin{minipage}[t]{0.49\textwidth}
        \centering
        \small
    \renewcommand{\arraystretch}{0.8}
        
        \resizebox{0.95\textwidth}{!}{
        \begin{tabular}{ccc}
        \toprule
        Modality & Top1 Acc$\uparrow$ & Top5 Acc$\uparrow$ \\ \midrule
        Task Model & 51.67\scriptsize{$\pm0.12$} & 73.45\scriptsize{$\pm0.15$}  \\ 
        V & 52.13\scriptsize{$\pm0.14$} & 74.25\scriptsize{$\pm0.18$}  \\
        V+T & 52.45\scriptsize{$\pm0.11$} & 75.06\scriptsize{$\pm0.13$}  \\ 
        \rowcolor{gray!20}
        \textbf{A+V+T} & \textbf{53.36\scriptsize{$\pm0.21$}}  &  \textbf{83.42\scriptsize{$\pm0.15$}}\\ 
        \bottomrule
        \end{tabular}
        }
        \vspace{-0.5em}
        \caption{\textcolor{black}{Multi-modal Input on Zero-shot Recognition Task.}}
        \vspace{-1.5em}
        \label{tab:class_abl}
    \end{minipage}
    \hfill
    \begin{minipage}[t]{0.49\textwidth}
        \centering
        \small
    \renewcommand{\arraystretch}{0.8}
        
        \resizebox{1\textwidth}{!}{
        \begin{tabular}{ccc} \toprule
        Modality & \begin{tabular}[c]{@{}c@{}}Frame\\ Consistency$\uparrow$\end{tabular} & \begin{tabular}[c]{@{}c@{}}Textual\\ alignment$\uparrow$\end{tabular}    \\ \midrule
        Task Model & 94.52\scriptsize{$\pm0.16$} & 29.53\scriptsize{$\pm0.21$}\\ 
        V & 95.82\scriptsize{$\pm0.13$} &  33.11\scriptsize{$\pm0.18$} \\ 
        V+T& 96.21\scriptsize{$\pm0.14$} & 33.86\scriptsize{$\pm0.20$} \\
        \rowcolor{gray!20} \textbf{V+T+D} & \textbf{96.56\scriptsize{$\pm0.15$}} & \textbf{34.01\scriptsize{$\pm0.25$}} \\ 
        \bottomrule
        \end{tabular}
        }
        \vspace{-0.5em}
        \caption{\textcolor{black}{Multi-modal Input on Text-to-Video Editing.}}
        \vspace{-1.5em}
        \label{tab:davis_edit_abl}
    \end{minipage}
\end{table}
\begin{table}[!]
    \centering
    \begin{minipage}[t]{0.49\textwidth}
        \centering
        \small
        \setlength{\tabcolsep}{0.5mm}
        \resizebox{0.95\textwidth}{!}{
        \begin{tabular}{ccccc}
        \toprule
        Modality & AJ$\uparrow$ & $<\delta^x_\textrm{avg}$$\uparrow$ & OA$\uparrow$ & TC$\downarrow$  \\ \midrule
        Task Model & 51.7\scriptsize{$\pm0.2$} & 67.5\scriptsize{$\pm0.2$} & 85.3\scriptsize{$\pm0.1$} & 0.74\scriptsize{$\pm0.02$} \\ 
        V & 52.3\scriptsize{$\pm0.1$} & 68.0\scriptsize{$\pm0.2$} & 85.6\scriptsize{$\pm0.2$} & 0.73\scriptsize{$\pm0.01$} \\ 
        V+T & 53.6\scriptsize{$\pm0.2$} & 68.5\scriptsize{$\pm0.1$} & 86.1\scriptsize{$\pm0.1$} & 0.72\scriptsize{$\pm0.02$} \\ 
        \rowcolor{gray!20} \textbf{V+T+D} & \textbf{54.4\scriptsize{$\pm0.2$}} & \textbf{68.8\scriptsize{$\pm0.1$}} & \textbf{86.6\scriptsize{$\pm0.2$}} & \textbf{0.70\scriptsize{$\pm0.01$}} \\ 
        \bottomrule
        \end{tabular}
        }
        \vspace{-0.5em}
        \caption{\textcolor{black}{Multi-modal Input on DAVIS pixel-tracking task.}}
        \vspace{-1.5em}
        \label{tab:tracking_abl}
    \end{minipage}
    \hfill
    \begin{minipage}[t]{0.49\textwidth}
        \centering
        \small
        \setlength{\tabcolsep}{1mm}
        \resizebox{1\textwidth}{!}{
        \begin{tabular}{cccccc} 
        \toprule
        Modality & B$\uparrow$ & M$\uparrow$ & C$\uparrow$ & S$\uparrow$ & mIoU$\uparrow$ \\  \midrule
        Task Model & 51.6\scriptsize{$\pm0.2$} & 38.4\scriptsize{$\pm0.1$} & 310.1\scriptsize{$\pm0.2$} & 54.5\scriptsize{$\pm0.2$} & 64.2\scriptsize{$\pm0.2$} \\ 
        V & 52.1\scriptsize{$\pm0.1$} & 38.4\scriptsize{$\pm0.2$} & 315.2\scriptsize{$\pm0.2$} & 55.0\scriptsize{$\pm0.1$} & 65.6\scriptsize{$\pm0.1$} \\ 
        V+T & 52.6\scriptsize{$\pm0.2$} & 38.7\scriptsize{$\pm0.1$} & 319.9\scriptsize{$\pm0.2$} & 55.2\scriptsize{$\pm0.1$} & 66.3\scriptsize{$\pm0.2$} \\ 
        \rowcolor{gray!20}
        \textbf{V+T+D} & \textbf{52.9\scriptsize{$\pm0.2$}} & \textbf{39.0\scriptsize{$\pm0.1$}} & \textbf{325.5\scriptsize{$\pm0.3$}} & \textbf{55.7\scriptsize{$\pm0.2$}} & \textbf{66.9\scriptsize{$\pm0.2$}} \\ \bottomrule
        \end{tabular}
        }
        \vspace{-0.5em}
        \caption{\textcolor{black}{Multi-modal Input on Drama Dataset.}}
        \vspace{-1.5em}
        \label{tab:drama_abl}
    \end{minipage}
\end{table}

\textbf{Multi-modal Input Ablation.}
\textcolor{black}{
We evaluate the impact of input modalities across Tables~\ref{tab:class_abl}--\ref{tab:drama_abl}.
The results show consistent performance gains across all tasks as additional modalities are integrated.
For instance, supplementing the video baseline with auxiliary modalities (e.g., text, depth, or audio) yields notable improvements in both Zero-shot Recognition (Top-1 Acc) and Pixel-level Tracking (AJ).
This confirms that SEN effectively leverages complementary multimodal cues to enhance robust video understanding.}

% 这个表格增加一些东东。

\begin{figure*}[t]
\begin{center}
\includegraphics[width=\linewidth]{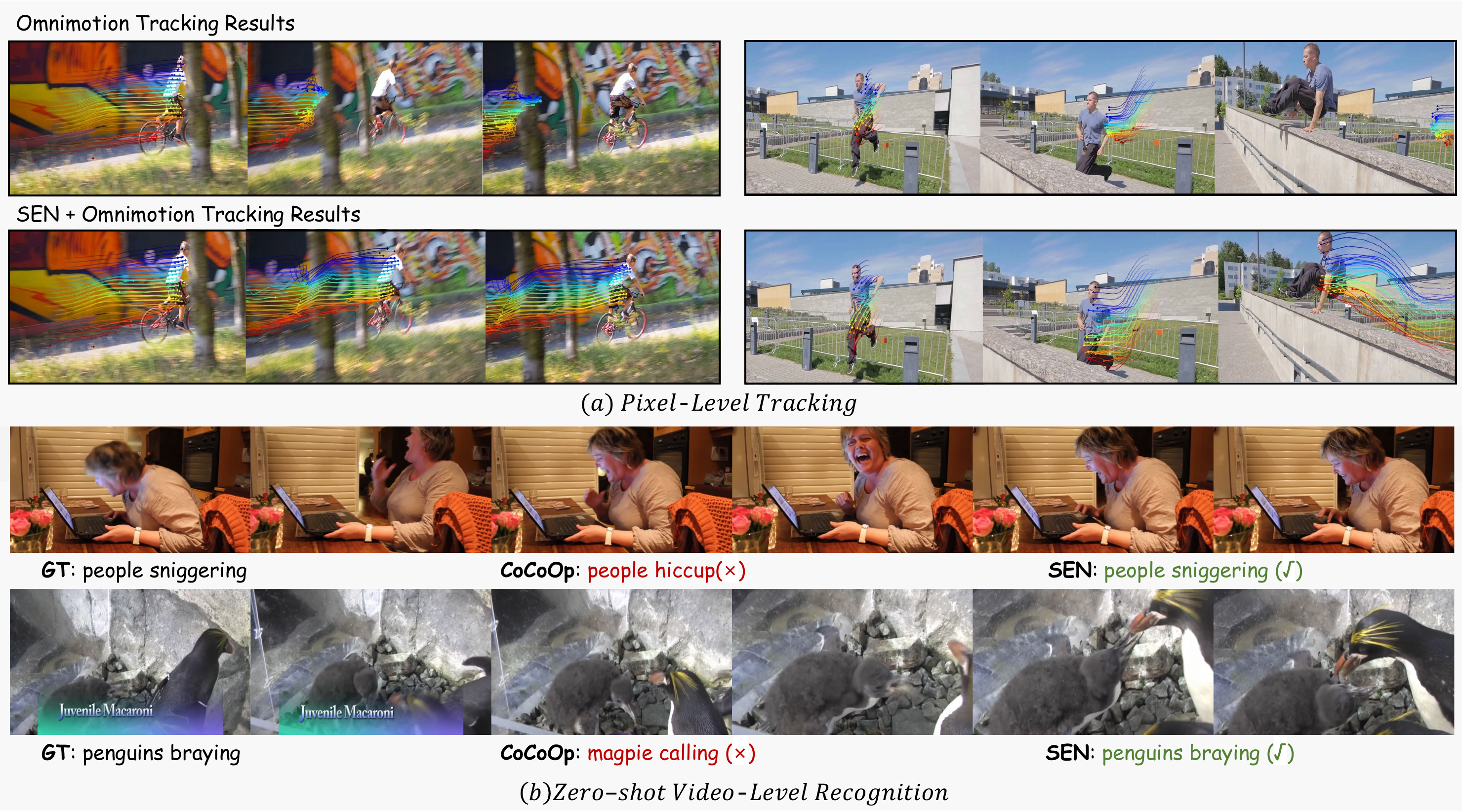}
\vspace{-1.5em}
\caption{Visualization of SEN. (a) is for the pixel-level tracking task. (b) is for the video-level zero-shot recognition task.}
\label{fig:tracking}
\end{center}
\vspace{-2em}
\end{figure*}

\textbf{Additional Trainable Params and Eval Time:}
As shown in Table~\ref{tab: trainable}, SEN is highly lightweight, introducing only 5-6M trainable parameters. 
The inference pipeline consists of multimodal pre-processing and SEN inference.
For pre-processing, depth estimation
(GLPN~\cite{kim2022global}) and caption generation (GPT-4o~\cite{gpt4}, using 4 sampled frames) take approximately
0.116s and 0.4s, 
\begin{wraptable}{r}{0.55\linewidth}
    \small
    \renewcommand{\arraystretch}{0.6}
    \setlength{\tabcolsep}{1mm}
    \centering
    \vspace {-1em}
    \resizebox{1\linewidth}{!}{
    \begin{tabular}{ccccc} \toprule
        Task & \begin{tabular}[c]{@{}c@{}}Pixel-level\\ Tracking \end{tabular}  & \begin{tabular}[c]{@{}c@{}}Zero-shot\\ Recognition \end{tabular}   & \begin{tabular}[c]{@{}c@{}}Automatic\\ Driving Chatting \end{tabular}  & \begin{tabular}[c]{@{}c@{}}One-shot \\ Video Editing\end{tabular}   \\  \midrule
        Modality & V+T+D & V+T+A & V+T+D & V+T+D \\  \midrule
        \begin{tabular}[c]{@{}c@{}}Foundation\\  Model\end{tabular}  & ImageBind &   VALOR   & ImageBind & ImageBind \\  \midrule
        \begin{tabular}[c]{@{}c@{}}Addition \\ Params(M)\end{tabular}    & 5.24 & 6.58 & 5.11 & 5.24 \\ \midrule
        \rowcolor{gray!20} \begin{tabular}[c]{@{}c@{}}Baseline\\ Eval Time\end{tabular} &  3.885s & 0.225s & 3.302s & 31.808s \\ \midrule
        \begin{tabular}[c]{@{}c@{}}Multi-modal\\ extraction Time\end{tabular} &  0.516s & 0s & 0.516s & 0.516s \\ \midrule
        \begin{tabular}[c]{@{}c@{}}SEN Inference\\ Time\end{tabular} &  0.346s & 0.178s & 0.346s & 0.346s \\ \midrule
        \rowcolor{gray!20}  \begin{tabular}[c]{@{}c@{}}Total\\ Eval Time\end{tabular} &  \begin{tabular}[c]{@{}c@{}}4.747s\\ \textcolor{black}{+0.862s}\end{tabular} & \begin{tabular}[c]{@{}c@{}}0.403s\\\textcolor{black}{+0.178s}\end{tabular}  & \begin{tabular}[c]{@{}c@{}}4.164s\\ \textcolor{black}{+0.862s}\end{tabular}  & \begin{tabular}[c]{@{}c@{}}32.670s\\ \textcolor{black}{+0.862s}\end{tabular}\\ \midrule
    \end{tabular}
    }
    \vspace{-1em}
    \caption{Additional Trainable Params and Eval Time.}
    \label{tab: trainable}
    \vspace {-2em}
\end{wraptable}
respectively. Note that tasks like VGGSound do not 
require this step. 
In inference, SEN operates
efficiently on 4 RTX A6000 GPUs, requiring merely 0.516s for ImageBind-based tasks and 0.178s for
VALOR-based tasks.
Even with pre-processing included, 
the total time per sample is $\approx$ 0.862s, demonstrating that SEN achieves an optimal balance between computational cost and performance.

\begin{figure*}
\begin{center}
\vspace{-2em}
\includegraphics[width=\linewidth]{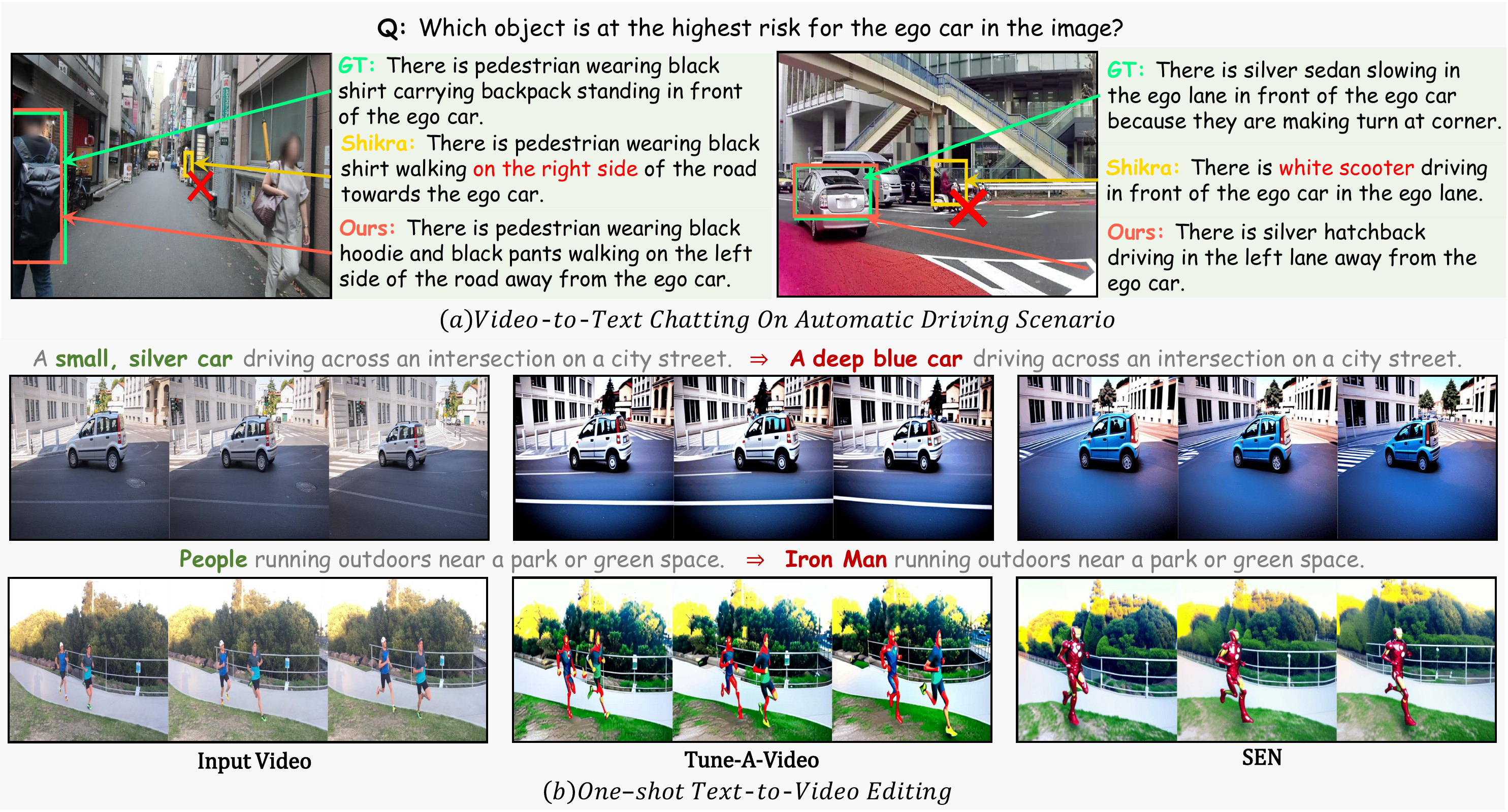}
\vspace{-2em}
\caption{Visualization of SEN. (a) is for the video-to-text chatting task for the autopilot scenario. (b) is for the text-to-video editing task.}
\label{fig:gener}
\end{center}
\vspace{-2em}
\end{figure*}

% In the pixel-level tracking task as shown in Fig~\ref{fig:tracking}(a),
% Omnimotion fails to maintain tracking information when a person is moving at high speed. In contrast, 

% TODO:增加这部分的细节，解释一下tsne的内容。
\textbf{Visualization.} 
\begin{figure*}[t]
\centering
\includegraphics[width=\textwidth]{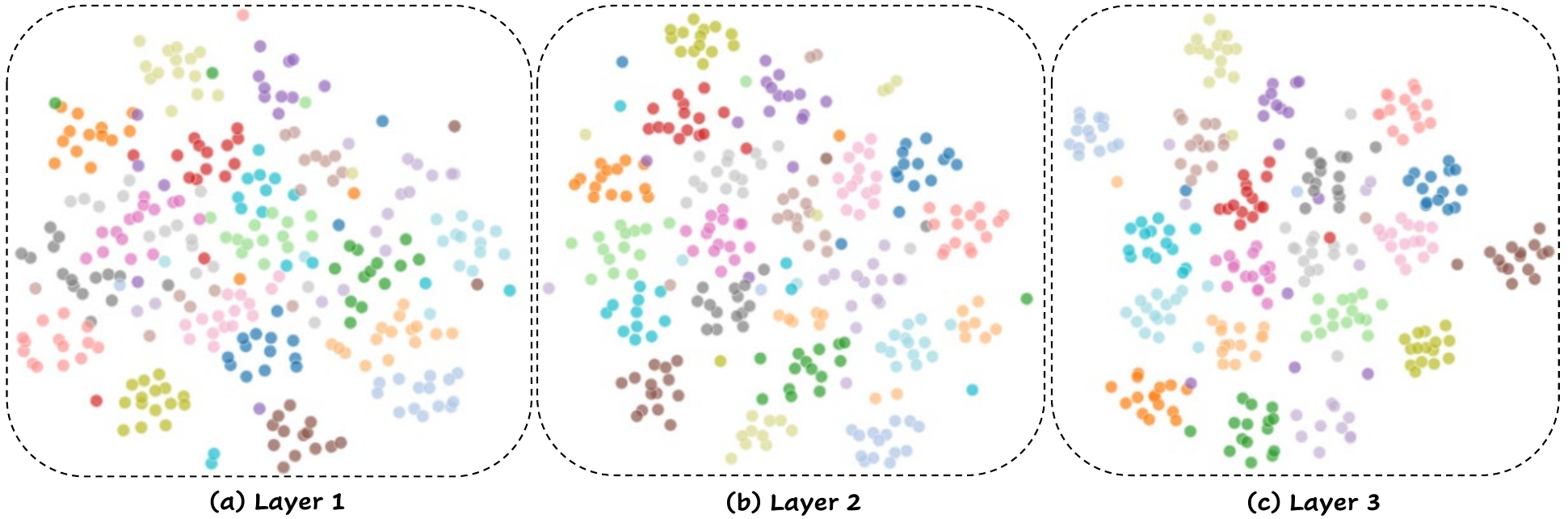}
\vspace {-1.5em}
\caption{\textcolor{black}{t-SNE visualization of the feature representations after different layers of the RA-encoder. We selected 20 classes from VGGSound with 15 samples each. From (a) to (c), the distinctiveness of the clusters improves significantly, demonstrating the effectiveness of the layer-wise processing.}}
\label{fig:tsne}
\vspace {-1em}
\end{figure*}
\textcolor{black}{
We provide qualitative comparisons in Fig.~\ref{fig:tracking}, Fig.~\ref{fig:tsne} and Fig.~\ref{fig:gener}. 
In Pixel-level Tracking} 
\textcolor{black}{(Fig.~\ref{fig:tracking}(a)) and Zero-shot Recognition (Fig.~\ref{fig:tracking}(b)), SEN demonstrates robust trajectory maintenance and discriminative power in ambiguous scenes.
For the Video-to-Text Chatting task (Fig.~\ref{fig:gener}(a)), our method accurately localizes the risk object and generates correct descriptions, effectively reducing semantic hallucinations.
In Text-to-Video Editing (Fig.~\ref{fig:gener}(b)), SEN achieves precise prompt alignment (e.g., Iron Man) while preserving temporal consistency.
Additionally, Fig.~\ref{fig:tsne} visualizes the feature evolution within the RA block using t-SNE. 
We constructed a diverse subset from the VGGSound dataset, consisting of 20 distinct categories to ensure visual clarity. For each category, we randomly selected 15 samples.
We extracted the prompt features $\mathbf{P}^{(i)}_{m}$ from the output of the 1st, 2nd, and 3rd layers of the RA-encoder. For each sample, the features were concatenated and projected into a 2D space using t-SNE.
As the layers deepen (from (a) to (c)), the feature distribution transitions from an entangled state to a clear cluster structure with well-defined boundaries. }
\begin{wrapfigure}{r}{0.55\linewidth}
% 图形内容
\vspace{-2.5em}
\begin{center}
\includegraphics[width=\linewidth]{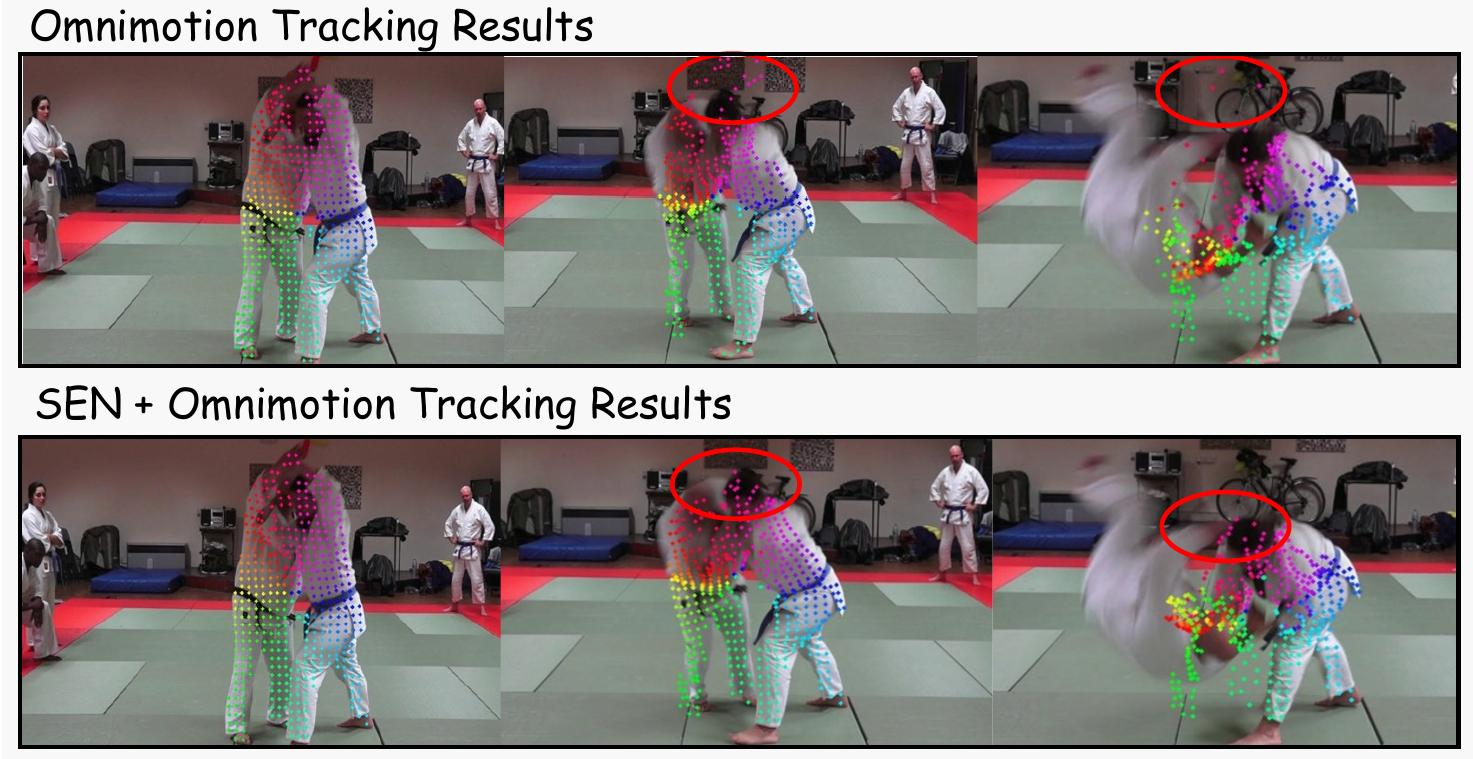}
\vspace{-2.5em}
\caption{Visualization of failure case.}
\label{fig:vis_wrong}
\end{center}
\vspace{-2.8em}
\end{wrapfigure} 
\textcolor{black}{This progression empirically verifies that our recursive mechanism progressively refines multimodal representations, yielding more discriminative features through deep interaction.
We also analyze a failure case in Fig.~\ref{fig:vis_wrong}.
While tracking rapidly moving and spinning targets remains challenging, SEN still exhibits superior stability compared to the baseline, maintaining better identity consistency even in failure scenarios.}

\textbf{Limitations and Future Work.}
\textcolor{black}{
Although SEN demonstrates remarkable generalization across diverse tasks, due to resource constraints, we prioritized two representative foundation models (ImageBind and VALOR). 
Future work will focus on expanding SEN to a broader spectrum of foundation encoders and incorporating additional modalities (e.g., thermal or IMU) to further verify the scalability of our recursive association paradigm.}

\section{Conclusion}
% need change
In conclusion, we propose a brand-new Super Encoding Network (SEN) for multimodal video understanding. 
SEN deeply \textcolor{black}{integrates} the multimodal knowledge of frozen foundation models.
This \textcolor{black}{parameter-efficient} structure \textcolor{black}{(requiring only $\sim$5.24M trainable parameters)} allows our SEN to \textcolor{black}{augment existing alignment with substantial Interaction Depth} and prompt various video understanding tasks in the downstream.
The proposed RA blocks progressively associate foundation models for deeper multimodal interaction via knowledge integrating, distributing, and prompting video-relevant knowledge in a recursive manner.
Extensive experiments show that SEN performs remarkably on four representative video tasks, from video tracking to recognition, from video chatting to editing, \textcolor{black}{validating its architectural robustness and parameter effectiveness}.

\bibliographystyle{elsarticle-num}
\bibliography{cas-refs} 
\end{document}